\documentclass[11pt]{article}
\pdfoutput=1
\usepackage[T1]{fontenc}
\usepackage[utf8]{inputenc}
\usepackage{lmodern}
\usepackage{textcomp}
\usepackage[a4paper,margin=2.5cm]{geometry}
\usepackage{graphicx}
\usepackage{longtable,booktabs,array,calc}
\usepackage{amsmath,amssymb}
\usepackage{microtype}
\usepackage[hidelinks]{hyperref}
\usepackage{etoolbox}
\providecommand{\tightlist}{\setlength{\itemsep}{0pt}\setlength{\parskip}{0pt}}

\providecommand{\real}[1]{#1}
\DeclareUnicodeCharacter{2212}{\ensuremath{-}}
\DeclareUnicodeCharacter{00D7}{\ensuremath{\times}}
\DeclareUnicodeCharacter{00B7}{\textperiodcentered}
\title{Exposure is not manifestation: measurement target and output resolution jointly determine which behavioural-faithfulness evaluator wins}
\author{Kwan Soo Shin\\[2pt]\normalsize PolymathMinds Lab, Asan, Republic of Korea\\
\normalsize sshin@pmminds.ai \textperiodcentered{} ORCID 0009-0001-5799-7556}
\date{}
\begin{document}
\maketitle
\begin{abstract}
Behavioural auditing asks whether a language model behaves as it claims, but detection scores are reported without separating two targets: whether a reply was produced under a behaviour-inducing condition (exposure) and whether the behaviour surfaced in it (manifestation). Scoring a compact 146-million-parameter auditor's frozen-representation read-out and a frontier judge against each label on the identical 720 replies, the gap between the instruments moves by roughly 0.2 AUROC when the target changes. Under the judge's deployed interface, a single verdict, the ranking reverses: the auditor leads on exposure, 0.804 against 0.718, and trails on manifestation, 0.690 against 0.811. Matching the output resolution from either direction, by asking the judge a target-specific question answered with a continuous confidence score or by thresholding the auditor's read-out, removes the reversal but not the interaction, which excludes zero at all three resolutions (0.207, 0.237 and 0.169). The target governs how far apart the instruments are; the interface governs whether that distance changes their order. The auditor's hyperbolic geometry confers no advantage here. A single behavioural-detection AUROC is under-specified: such claims are comparable only when they state the estimand, the evaluator, and its output interface.
\end{abstract}
\subsection{1. Introduction}\label{introduction}

Behavioural evaluation of language models rests today on two instruments, human raters and frontier-model judges{[}1,2{]}, and reports their agreement as a single detection score. That score hides a question the field has not asked: what, exactly, is being detected. When an evaluation induces a behaviour through a prompt and then measures whether an instrument can flag it, two different targets are entangled. One is exposure: was this reply produced under the behaviour-inducing condition. The other is manifestation: did the behaviour actually surface in the reply. The two are not the same, because an inducing prompt does not always produce the behaviour; in the evaluation we report, only 61 percent of trait-conditioned replies manifested the trait.

We show that the distinction is decisive, and we separate it from a second factor with which it is easily confused. Scoring a compact 146-million-parameter auditor's frozen-representation read-out and a frontier judge against each label on the identical replies, the distance between the two instruments moves by roughly 0.2 AUROC when the target changes. That movement survives every output format we test. What does not survive is the ranking. Under the judge's deployed interface, a single verdict, the auditor leads on exposure and trails on manifestation, so the ordering reverses. Matching the output resolution from either direction, by asking the judge a target-specific question answered with a continuous confidence score, or by thresholding the auditor's read-out at 0.5 without data-dependent optimisation, removes the reversal: the two instruments become statistically indistinguishable on exposure, and the judge leads decisively on manifestation. Yet the interaction, the change in the gap between the two targets, excludes zero at all three resolutions. The target governs how far apart the instruments are; the interface governs whether that distance is enough to change their order.

We report every cell of this design because reporting only one would reproduce the error the paper is about. A single AUROC for behavioural detection is under-specified in two ways at once: it does not say which target was scored, and it does not say what the evaluator was permitted to emit. A compact representation carries the structural fingerprint of the inducing condition, while a semantic judge reads whether the content realises the behaviour; which of those is more useful depends on the question, and which appears better depends additionally on the answer format. We are equally deliberate about the substrate. The auditor's representation is hyperbolic, but for this task the curvature confers no advantage; a Euclidean read-out on the identical frozen representation is better. We report this boundary because it locates the transferable signal in a purpose-built representation and its supervision, not in geometry or scale. The contribution is a measurement principle: behavioural-faithfulness claims must state the estimand and the evaluator together, and evaluation is better served by a panel whose members fail on different items than by any single judge.

\subsection{2. Related work}\label{related-work}

This paper is about what a detection score denotes, so its home is the measurement literature rather than the behaviour literature. We read 93 contributions against two questions: what each established, and where it stopped. Eleven lineages converge on the same unoccupied position.

\textbf{Measurement validity became a first-class question, and stopped one step short of plurality.} Evaluation research now asks whether a score measures the construct it names rather than assuming it does{[}3,4,5{]}. The case has been made empirically: a survey of 445 language-model benchmarks found that the phenomena, tasks, and scoring rules in common use undermine the claims drawn from them{[}6{]}; leaderboards are shown to be shaped by participation and reporting choices as much as by capability{[}7{]}; benchmark composition carries the assumptions of the communities that build it{[}8,9{]}; and scores move with incidental properties of the input such as position in context{[}10{]}. In response, conditions for construct validity have been derived from psychometric theory{[}11,12{]}, evaluation has been reframed as a social-science measurement problem{[}13{]}, validity-centred frameworks have been proposed{[}14{]}, and the field has been called toward an evaluation science{[}15,16,17{]}. This lineage has done the difficult work of asking whether a number means what it is taken to mean, but its unit of analysis is one score against one construct. The case where one score aggregates two distinct targets, so that separating them changes the comparison of instruments, is not in the frame. Validity is asked about; plurality of estimand is not.

\textbf{Human disagreement was reclassified as signal, and located in the rater.} Agreement statistics entered computational linguistics as a discipline of their own{[}18,19,20,21,22{]}, and a substantial line then argued that annotator disagreement is not noise to be averaged away but information about the item and the annotator{[}23,24,25,26,27{]}, that identity and perspective shape the label{[}28,29{]}, that crowdsourced judgement of open-ended generation is unreliable{[}30{]}, and that representative participation changes what the label records{[}31{]}. This lineage made disagreement respectable and located the variation in who is judging. It did not consider two labels on the same reply that are each correct because each answers a different question, which is the case this paper reports.

\textbf{Linear read-outs of frozen representations became a mature instrument, disciplined from the representation side only.} Probing began with linear classifiers on intermediate layers{[}32{]} and acquired its central safeguard when control tasks separated probe capacity from representation content{[}33{]}; the paradigm was then examined for what probing accuracy licenses{[}34,35,36{]} and extended by amnesic counterfactuals{[}37{]} and adversarial concept erasure{[}38{]}. A more recent line reads structured content directly out of frozen activations: latent knowledge without supervision{[}39{]} and from models trained to be unreliable{[}40{]}, truth as an emergent linear direction{[}41{]}, space and time as recoverable coordinates{[}42{]}, linear world-model features{[}43{]}, and top-down representation engineering{[}44{]} with inference-time intervention{[}45{]}. Every methodological safeguard in this lineage guards one error, concluding that a representation encodes something the probe in fact supplied. The complementary error, leaving unspecified what the probe's target denotes, has no safeguard, because the target has been treated as given.

\textbf{Stated behaviour and enacted behaviour were shown to diverge, and the divergence was compressed into one number.} Chain-of-thought explanations were shown not to track the computation that produced the answer{[}46{]}, faithfulness became a measurable quantity{[}47{]}, and the measurement was extended to counterfactual simulatability{[}48{]} and separated from mere plausibility{[}49{]}; generated content was surveyed for confabulation{[}50{]} and instruction-following was compared against human performance{[}51{]}. This is the lineage nearest to ours in motivation. Its measurements, however, are reported as one faithfulness figure per system. The condition under which the figure was obtained does not survive into the number, so two such figures cannot be compared unless the conditions happened to match.

\textbf{Judges are audited for their own bias, never for ambiguity in the label.} The frontier judge became the default evaluator{[}2{]} and its failure modes were mapped quickly: preference for its own generations{[}52{]}, likelihood-driven bias{[}53{]}, and length effects strong enough to require explicit correction{[}54{]}. Each of these locates the defect in the judge. None asks whether a judge and a competing instrument might diverge because the label they are scored against admits two readings.

\textbf{Behaviour is induced, and the inducing condition is used as the label.} Model behaviour can be elicited and observed at scale{[}1,55{]}, and the elicited behaviours are known to be products of preference optimisation{[}56{]}, with sycophancy characterised directly{[}57{]} and partially mitigated{[}58{]}; alignment properties are known to erode under fine-tuning{[}59{]}, and safety benchmarks have been shown to track capability rather than safety{[}60{]}. Persona and role instructions move behaviour systematically{[}61,62,63,64{]} and destabilise it across a conversation{[}65,66{]}. In this design a reply produced under a trait-inducing prompt is a positive instance whether or not the trait appeared in it. One recent study begins to separate the two, finding that self-report and behaviour cohere only selectively and that coherence collapses where behaviour is strongly primed by context, as with sycophancy{[}67{]}. That result names the gap but does not convert it into two measurement targets scored on the same items.

\textbf{Three lineages bound the study rather than motivate it.} Compact specialised models can outperform much larger generalists{[}68,69,70,71,72,73{]}, which is what makes the comparison here worth running, though that literature adjudicates on a single task score. Hyperbolic representation is well founded for hierarchical structure{[}74,75,76,77,78,79{]}, which motivates the auditor's substrate without predicting whether curvature helps behavioural detection. Generalisation to unseen groups requires explicit design rather than random splits{[}80,81,82,83,84,85{]}, which is why the evaluation is leave-one-generator-out; that literature specifies how to hold out, not what the held-out label denotes. Finally, the case for independent behavioural audit of deployed systems is established{[}86,87,88,89,90,91{]}, and it supplies the stakes: an audit regime must choose instruments and currently has no principled basis for choosing.

\textbf{Relation to the author's companion work.} A separate manuscript by the author, under review elsewhere and disclosed in the cover letter, concerns process compliance: whether a system that states it will follow an instructed procedure in fact follows it, measured against tool-call logs. That work asks whether a dimension of behaviour has been measured at all. The present paper asks a different question about a dimension already being measured, namely what a detection score on it denotes, and answers it by showing that one score conceals two targets whose separation reverses the ranking of instruments. The two share subject matter, part of a bibliography, and one asset, disclosed here and in the cover letter: the ten-rater human-rating study of Section 3.5, which is reported there as a detectability result and serves here as the text-layer baseline condition. They share no other experiment or dataset, none of the present paper's display items or central results appears there, and neither paper's conclusions depend on the other.

\textbf{The empty position.} Arrange the field on two axes, the label a study scores against and the instrument it scores with. Human raters, frontier judges, representation read-outs, and compact specialised models all appear; both labels appear. What does not appear is a study that scores both labels on the identical replies and asks what the change of target does to the comparison of instruments. Eleven lineages approach that position from different directions, and each, as set out above, stops one step short. This paper occupies the position. Scoring exposure and manifestation on the same 720 replies, the distance between a compact auditor and a frontier judge moves by roughly 0.2 AUROC with the target, at every output resolution tested, and under the deployed interface that movement reverses their ranking. A single behavioural-detection AUROC is therefore under-specified, and such a claim is comparable with another only when it states its estimand and its instrument together.

\subsection{3. Results}\label{results}

\subsubsection{3.1 A compact behavioural auditor and its in-distribution faithfulness}\label{a-compact-behavioural-auditor-and-its-in-distribution-faithfulness}

The auditor is a decoder-only transformer of 146 million parameters{[}70{]} (depth 16, width 512, 128-dimensional hyperbolic embedding), trained from scratch on 812,833 paired utterance-and-trace examples with 90,314 held out. Each example pairs an utterance, a behavioural trace elicited through a preference-neutralising disclosure protocol{[}56{]}, and a consistency label, and the auditor predicts from the utterance alone whether the trace would be consistent if revealed{[}46{]}. A shared encoder is read out through fourteen classification heads.

All numbers below come from a single fixed checkpoint (SHA-256 prefix 3127f6c7), re-evaluated end to end so that the faithfulness result and the trait-detection results share one model. On the held-out set the faithfulness head reaches 90.7 percent binary accuracy over its 25,064 labelled items (verdict head 87.4 percent over 36,008 items; fourteen-head macro F1 0.799; Table 1). This is an in-distribution held-out result: the 90,314-item set is held out at the item level and is not stratified by generator family, so it speaks to faithfulness detection on unseen items, not to cross-family generalization, which we test separately in Section 3.2. On a 32-session stratified sample with automatic tool-log ground truth, the auditor's refuse head agrees with the gold at Cohen kappa{[}20{]} 0.566 (accuracy 81.2 percent), matching frontier zero-shot judges at that sample size (GPT-4o kappa 0.66, Claude-Opus kappa 0.42, overlapping intervals at n = 32).

\subsubsection{3.2 The measurement target moves the gap; the output interface moves the ranking}\label{the-measurement-target-moves-the-gap-the-output-interface-moves-the-ranking}

The cross-family test uses a leakage-controlled companion-trait evaluation: three trait conditions (sycophancy{[}57{]}, dependence-fostering, confabulated memories{[}50{]}) against matched neutral conditions, generated as style-controlled pairs that share persona, probe, memory context, and generator within each pair, across three generator models from two providers, for 720 replies. Shared generator and context within each pair rule out style separation, and leave-one-generator-out evaluation rules out generator fingerprints.

Each reply carries two labels. The exposure label marks whether the reply was produced under a trait-inducing prompt. The manifestation label, assigned at data-construction time by an independent frontier judge, marks whether the trait actually surfaced; by this label 61.4 percent of trait-conditioned replies manifested the trait. We score both the auditor and a frontier zero-shot judge against each label in turn, on the identical replies.

The comparison is run at three output resolutions, because a judge that returns one verdict and a read-out that returns a continuous score are not the same instrument (Table 2). Under the judge's deployed interface, a single YES or NO verdict, the ranking reverses. On exposure, a logistic read-out{[}33{]} of the auditor's frozen representation transfers across unseen families at mean AUROC 0.804, exceeding the frontier judge's 0.718; the paired difference is 0.086 (95 percent interval 0.052 to 0.120). On manifestation, measured on the same replies, the frontier judge leads at 0.811 against the auditor's 0.690; the paired difference is negative 0.121 (negative 0.193 to negative 0.050). Both differences exclude zero. Intervals throughout are cluster bootstrap intervals that resample the 360 trait-and-neutral pairs, rather than individual replies, within each held-out generator, so that the paired construction of the evaluation set is carried into the uncertainty.

Two further resolutions remove the difference in output format, approaching it from opposite sides. In the resolution-matched control the judge is asked a target-specific question, one for exposure and one for manifestation, and returns a continuous confidence score, the normalised probability of its YES token, rather than a verdict, matching the auditor's continuous read-out; the judge then reaches 0.834 on exposure and 0.958 on manifestation. In the mirror-image control the auditor's read-out is thresholded at 0.5, without data-dependent optimisation, and the two instruments are compared verdict against verdict; the auditor then reaches 0.736 on exposure and 0.660 on manifestation. Under both matchings the reversal disappears, and it disappears in the same way: on exposure the two instruments become statistically indistinguishable (negative 0.031, interval negative 0.065 to 0.005 in the continuous comparison; 0.018, negative 0.017 to 0.051 in the binary one), while on manifestation the frontier judge leads decisively (negative 0.268, negative 0.333 to negative 0.202; and negative 0.151, negative 0.220 to negative 0.080). The compact auditor holds its ground only on the target its representation was aligned to.

What survives all three matchings is the interaction. The change in the auditor-minus-judge gap between the two targets is 0.207 (0.140 to 0.276) under the deployed verdict interface, 0.237 (0.172 to 0.307) under the resolution-matched control, and 0.169 (0.102 to 0.237) under the binary comparison; every interval excludes zero. The measurement target therefore moves the distance between the two instruments by roughly 0.2 AUROC whatever the output format, while the output format decides whether that movement is large enough to change their order. We report the matched controls as primary results rather than robustness checks, because a comparison that leaves output resolution free cannot separate the two factors, and separating them is what this paper claims to do.

The pattern admits a legible reading. The auditor's frozen representation encodes a structural fingerprint of the inducing condition, which transfers across generators but overshoots when the condition did not produce the behaviour; the frontier judge reads semantic content, which tracks manifestation but does not carry a generator-invariant signature of the condition. This reading is an interpretation of the performance pattern; what is directly established is where the exposure signal lives. Two controls place it in the representation: the fourteen pre-trained heads do not transfer zero-shot to this task (all AUROC below 0.5), and fine-tuning the final decoder block on the exposure label reduces held-out AUROC to 0.729, so the frozen representation, not further training, carries the transferable signal.

Two cautions bound the manifestation half of the comparison, and we state them plainly. First, the auditor's representation was shaped by exposure-adjacent supervision while the judge is zero-shot, so the distances reflect instrument-target alignment as much as raw capability; they say the estimand moves the comparison, not that either instrument is intrinsically better. Second, the manifestation label is the output of a single automatic judge, so the manifestation metric is agreement with that judge rather than with human-verified behaviour, and the frontier judge's advantage on it is in part agreement between two frontier judges. A human-verified manifestation subset is the decisive next step, and until it exists we treat the manifestation ranking as established against an automatic reference, not against ground truth.

\subsubsection{3.3 The two instruments fail on different replies}\label{the-two-instruments-fail-on-different-replies}

Even where the auditor is the weaker instrument on manifestation, its errors do not coincide with the frontier judge's. Of the 225 replies the automatic judge labelled as manifesting, the frontier judge disagrees on 64; the auditor's read-out recovers 37 of those 64. Across all replies the two instruments misclassify at different marginal rates: a McNemar test{[}19{]} on the discordant pairs rejects equality of those rates (p = 1.07 × 10\textsuperscript{−20}, two-sided exact). That test bears on marginal rates only, and we use it only in that sense. The association between the two error events is a separate quantity and we report it separately: the two-by-two correctness table (both correct 484, auditor alone 39, judge alone 170, neither 27) gives an odds ratio of 1.97 and a phi coefficient of 0.097, so the errors are weakly positively associated rather than independent. It is the discordance, not independence, that a combination can exploit{[}92{]}.

Because 39 and 170 replies are recovered uniquely by one instrument or the other, two-instrument rules reach operating points neither instrument reaches alone (Table 3). The inclusive rule raises recall against the manifestation label from 0.716 for the judge alone to 0.880, at a false-positive rate of 0.253 against the judge's 0.004 and a precision of 0.613: a high-recall screen rather than a precise adjudicator. The exclusive rule is the opposite corner of the same discordance, with recall 0.507 at precision 1.000 and no false positives across 495 negatives: a high-precision confirmer. Neither rule is offered as an improvement in accuracy; both are offered as evidence that the two instruments carry non-redundant information about the same replies. We report this as a provisional result: it is computed against the automatic manifestation label, and its standing as a genuine coverage gain, rather than recovered agreement with the labelling judge, awaits the human-verified subset of Section 3.2. The per-family manifestation counts are uneven, with only 15 manifesting replies on the most distant generator, so per-family manifestation AUROCs there carry wide intervals and we report the mean over generators as primary.

\subsubsection{3.4 The geometry is a boundary, not the mechanism}\label{the-geometry-is-a-boundary-not-the-mechanism}

The auditor's representation is hyperbolic{[}74,75,77{]}, but the detection signal is not geometric. Under a matched read-out on the identical frozen representation, the hyperbolic Lorentz read-out is significantly worse than a Euclidean-pooled read-out for this task (paired difference in AUROC 0.108, 95 percent confidence interval 0.071 to 0.147, leave-one-generator-out; Figure 3). We therefore make no curvature claim for behavioural-faithfulness detection. The transferable capability is a property of a representation purpose-built and supervised for the task; the hyperbolic substrate's advantages, where they exist, concern hierarchical organisation and are a separate matter from the detection reported here.

\subsubsection{3.5 Human raters and the text-layer blind spot}\label{human-raters-and-the-text-layer-blind-spot}

The text-layer blind spot on this benchmark family is established by a blinded human evaluation reported in full in the earlier, broader version of this work{[}93{]}, whose rating data and analysis pipeline are included in the present deposit. Reading only text outputs, the panel reaches Fleiss kappa{[}21{]} 0.130 on session completion, slight agreement{[}22{]} practically indistinguishable from chance, and its majority identifies none of the sessions in which the audited model genuinely completed the behaviour, while the single rater who additionally consulted the underlying tool-call log was alone in approaching the true prevalence. We take that result as the baseline condition for the present work rather than re-deriving it: on this task the gap is invisible at the text layer, not merely difficult. Against that baseline the auditor's faithfulness head reaches 90.7 percent on the same benchmark family, reading what the text-only panel cannot.

The human evaluation and the trait-detection evaluation draw on the same benchmark family but are not co-registered item for item, use different label granularities, and report different statistics (agreement with gold versus inter-rater agreement). We present them side by side rather than as a paired comparison, and a co-registered comparison on a single prespecified set is scoped to follow-up.

The heads differ in accuracy and in sample size, and the two that carry the claims reported here are also the two evaluated on the largest held-out sets: the verdict head at 0.874 on 36,008 items and the faithfulness head at 0.907 on 25,064 (Figure 1, Table 1).

\begin{figure}[t]
\centering
\includegraphics[width=0.98\linewidth]{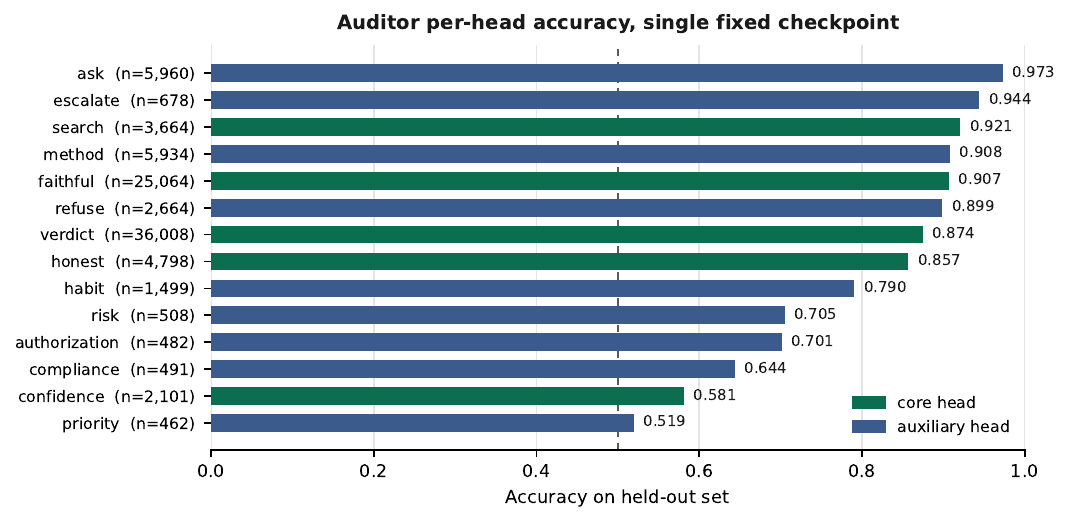}
\caption{\textbf{Auditor per-head accuracy on the 90,314-item held-out set.} Accuracy for each of the fourteen classification heads read out from the shared encoder, from a single fixed checkpoint (SHA-256 prefix 3127f6c7), with the number of held-out items for each head shown beside its name. Core heads, which carry the claims reported in the text, are distinguished from auxiliary heads. The dashed line marks chance for a binary head; heads with more than two classes have lower chance levels.}
\label{fig:1}
\end{figure}

\begin{table}[!htp]
\noindent \textbf{Table 1. Auditor per-head accuracy on the 90,314-item held-out set, single fixed checkpoint (SHA-256 prefix 3127f6c7; representative heads).}\par\medskip
\centering\small
\begin{tabular}{@{}lllll@{}}
\toprule\noalign{}
Head & Tier & N & Accuracy & Macro F1 \\
\midrule\noalign{}
verdict & core & 36,008 & 0.874 & 0.876 \\
faithful & core & 25,064 & 0.907 & 0.907 \\
honest & core & 4,798 & 0.857 & 0.857 \\
search & core & 3,664 & 0.921 & 0.921 \\
confidence & core & 2,101 & 0.581 & 0.554 \\
ask & aux & 5,960 & 0.973 & 0.973 \\
method & aux & 5,934 & 0.908 & 0.909 \\
refuse & aux & 2,664 & 0.899 & 0.898 \\
Overall (macro, 14 heads) & n/a & 90,313 & n/a & 0.799 \\
Core heads (5) & n/a & n/a & n/a & 0.823 \\
\bottomrule\noalign{}
\end{tabular}
\end{table}

Each held-out record carries exactly one head label; one record's label field is malformed in the source file and is excluded from every head, so head-level counts sum to 90,313 of the 90,314 records.

The estimand interaction is positive not only in the mean but in every held-out generator separately, at all three output resolutions (Figure 2, Table 2).

\begin{figure}[t]
\centering
\includegraphics[width=0.98\linewidth]{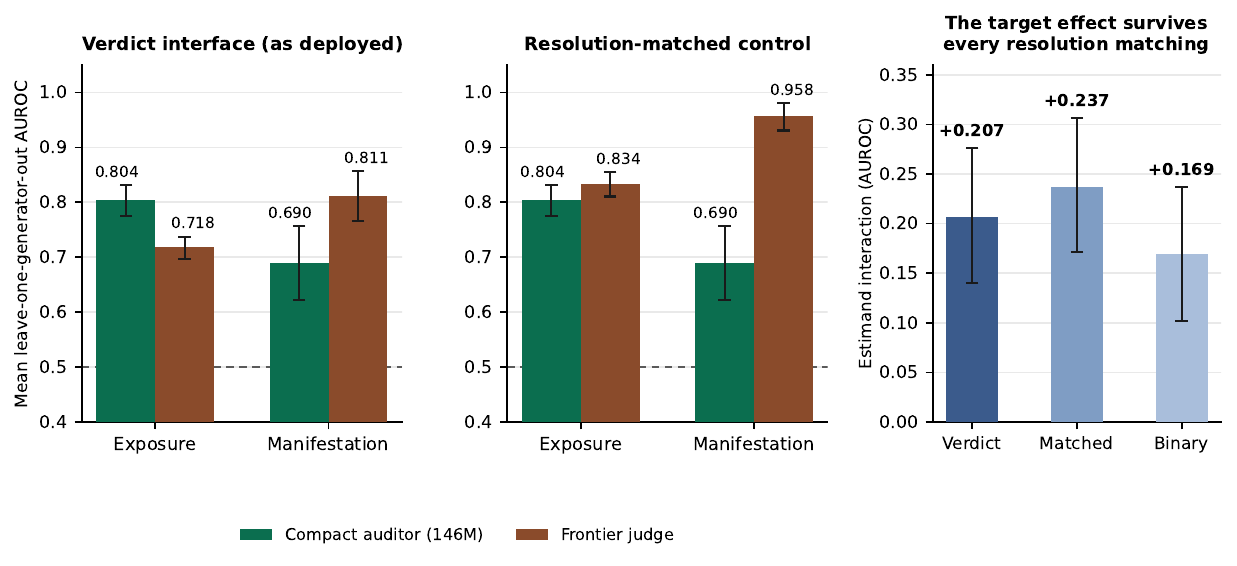}
\caption{\textbf{The measurement target moves the gap; the output interface moves the ranking.} Left, mean leave-one-generator-out AUROC for the compact auditor's linear read-out and for the frontier judge under the judge's deployed interface, a single verdict, scored against the exposure label and against the manifestation label on the identical replies; the ordering reverses between the two targets. Centre, the same comparison under the resolution-matched control in which the judge answers a target-specific question with a continuous confidence score; the judge now leads on both targets and the ordering does not reverse. Right, the estimand interaction, defined as the exposure gap minus the manifestation gap, at all three output resolutions, including the binary comparison in which the auditor's read-out is thresholded at 0.5; every interval excludes zero, so the target effect is not an artefact of the answer format. Error bars are 95 percent cluster bootstrap intervals resampling trait-and-neutral pairs within each held-out generator. Dashed lines mark chance.}
\label{fig:2}
\end{figure}

\begin{table}[!htp]
\noindent \textbf{Table 2. Companion-trait detection by estimand at three output resolutions, leave-one-generator-out (720 replies). Intervals are 95\% cluster bootstrap intervals resampling the 360 trait-and-neutral pairs within each held-out generator.}\par\medskip
\centering\small
\begin{tabular}{@{}
  >{\raggedright\arraybackslash}p{(\linewidth - 8\tabcolsep) * \real{0.2000}}
  >{\raggedright\arraybackslash}p{(\linewidth - 8\tabcolsep) * \real{0.2000}}
  >{\raggedright\arraybackslash}p{(\linewidth - 8\tabcolsep) * \real{0.2000}}
  >{\raggedright\arraybackslash}p{(\linewidth - 8\tabcolsep) * \real{0.2000}}
  >{\raggedright\arraybackslash}p{(\linewidth - 8\tabcolsep) * \real{0.2000}}@{}}
\toprule\noalign{}
\begin{minipage}[b]{\linewidth}\raggedright
Output resolution
\end{minipage} & \begin{minipage}[b]{\linewidth}\raggedright
Estimand
\end{minipage} & \begin{minipage}[b]{\linewidth}\raggedright
Auditor (146 M)
\end{minipage} & \begin{minipage}[b]{\linewidth}\raggedright
Frontier judge
\end{minipage} & \begin{minipage}[b]{\linewidth}\raggedright
Difference {[}95\% CI{]}
\end{minipage} \\
\midrule\noalign{}
Deployed verdict interface & Exposure & 0.804 & 0.718 & +0.086 {[}+0.052, +0.120{]} \\
continuous read-out vs single verdict & Manifestation & 0.690 & 0.811 & −0.121 {[}−0.193, −0.050{]} \\
& \textbf{Estimand interaction} & & & \textbf{+0.207 {[}+0.140, +0.276{]}} \\
Resolution-matched & Exposure & 0.804 & 0.834 & −0.031 {[}−0.065, +0.005{]} \\
continuous vs continuous, target-specific & Manifestation & 0.690 & 0.958 & −0.268 {[}−0.333, −0.202{]} \\
& \textbf{Estimand interaction} & & & \textbf{+0.237 {[}+0.172, +0.307{]}} \\
Binary & Exposure & 0.736 & 0.718 & +0.018 {[}−0.017, +0.051{]} \\
verdict vs verdict, threshold 0.5, no data-dependent optimisation & Manifestation & 0.660 & 0.811 & −0.151 {[}−0.220, −0.080{]} \\
& \textbf{Estimand interaction} & & & \textbf{+0.169 {[}+0.102, +0.237{]}} \\
Per family, exposure (continuous) & & 0.657 / 0.915 / 0.839 & 0.508 / 0.817 / 0.829 (verdict); 0.575 / 0.961 / 0.966 (matched) & \\
Per family, manifestation (continuous) & & 0.446 / 0.848 / 0.776 & 0.696 / 0.869 / 0.869 (verdict); 0.894 / 0.993 / 0.985 (matched) & \\
Pre-trained heads, zero-shot & Exposure & \textless{} 0.5 & n/a & \\
Final block fine-tuned & Exposure & 0.729 & n/a & \\
\bottomrule\noalign{}
\end{tabular}
\end{table}

The verdict interface is the judge as deployed: one question about manifestation, answered YES or NO. The resolution-matched interface asks a target-specific question and takes the normalised probability of the YES token, matching the auditor's continuous output; the binary comparison instead reduces the auditor to the judge's format. The estimand interaction is the difference between the exposure gap and the manifestation gap; it is of similar size at all three resolutions, while the sign of the exposure gap is not.

Per-family columns are claude-haiku / gpt-4o / gpt-4o-mini; the claude-haiku manifestation fold has 15 positive replies and a correspondingly wide interval.

\begin{figure}[t]
\centering
\includegraphics[width=0.85\linewidth]{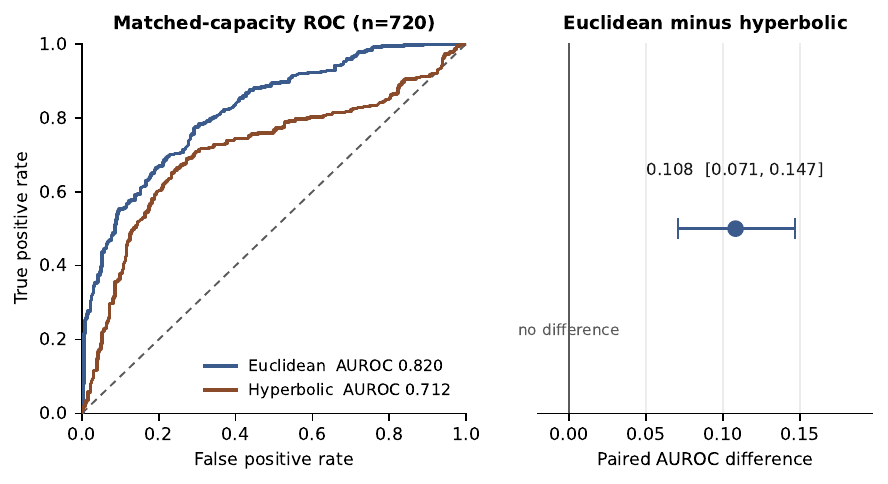}
\caption{\textbf{The geometry is a boundary, not the mechanism.} Left, receiver operating characteristic curves for a hyperbolic Lorentz read-out and a Euclidean-pooled read-out taken from the identical frozen representation under matched capacity (n = 720). Right, the paired difference in AUROC between the two read-outs with its 95 percent bootstrap confidence interval, which excludes zero. The transferable signal is a property of purpose-built representation and supervision rather than of curvature.}
\label{fig:3}
\end{figure}

\begin{table}[!htp]
\noindent \textbf{Table 3. Discordance under the manifestation label (225 manifesting and 495 non-manifesting replies); provisional, scored against the automatic manifestation label. Auditor thresholded at 0.5 without data-dependent optimisation.}\par\medskip
\centering\small
\begin{tabular}{@{}
  >{\raggedright\arraybackslash}p{(\linewidth - 8\tabcolsep) * \real{0.2000}}
  >{\raggedright\arraybackslash}p{(\linewidth - 8\tabcolsep) * \real{0.2000}}
  >{\raggedright\arraybackslash}p{(\linewidth - 8\tabcolsep) * \real{0.2000}}
  >{\raggedright\arraybackslash}p{(\linewidth - 8\tabcolsep) * \real{0.2000}}
  >{\raggedright\arraybackslash}p{(\linewidth - 8\tabcolsep) * \real{0.2000}}@{}}
\toprule\noalign{}
\begin{minipage}[b]{\linewidth}\raggedright
Metric
\end{minipage} & \begin{minipage}[b]{\linewidth}\raggedright
Frontier judge
\end{minipage} & \begin{minipage}[b]{\linewidth}\raggedright
Auditor (146 M)
\end{minipage} & \begin{minipage}[b]{\linewidth}\raggedright
Inclusive rule
\end{minipage} & \begin{minipage}[b]{\linewidth}\raggedright
Exclusive rule
\end{minipage} \\
\midrule\noalign{}
Recall vs manifestation label & 0.716 & 0.671 & 0.880 & 0.507 \\
Precision & 0.988 & 0.551 & 0.613 & 1.000 \\
Specificity & 0.996 & 0.752 & 0.747 & 1.000 \\
False-positive rate & 0.004 & 0.248 & 0.253 & 0.000 \\
Replies recovered from the judge's 64 disagreements & n/a & 37 & n/a & n/a \\
Correctness table vs judge (all 720) & both 484 & auditor alone 39 & judge alone 170 & neither 27 \\
Error association (odds ratio; phi) & 1.97; 0.097 & & & \\
McNemar on marginal error rates & p = 1.07 × 10\textsuperscript{−20} (two-sided exact) & & & \\
\bottomrule\noalign{}
\end{tabular}
\end{table}

\subsection{4. Discussion}\label{discussion}

The central result is a caution about measurement, and it has two parts that are usually reported as one. Behavioural auditing induces a behaviour, asks an instrument to detect it, and reports a single number; that number silently answers one of two different questions, and it silently fixes what the instrument was allowed to say. Separating the two shows that they do different work. The target sets the distance between instruments, moving the auditor-judge gap by about 0.2 AUROC at every output resolution we test. The interface sets whether that distance reorders them: under the deployed verdict interface the ordering reverses, and under matched resolutions, approached from either direction, it does not. The lesson is not that a small model beats or loses to a large one. Neither statement is well formed until the estimand is named, and a comparison that names the estimand but not the answer format can still be read backwards.

This has a constructive corollary. Because the two instruments read different things, a compact representation reading structure and a frontier judge reading content, their errors fall on different replies, and an evaluation panel that combines them covers more than either alone. We report that gain as provisional, because our manifestation labels come from a single automatic judge and the clean test is a human-verified subset, but the discordance itself is strong (McNemar p = 1.07 × 10\textsuperscript{−20}) and does not depend on which instrument is called better. The honest boundary on geometry belongs to the same argument: the transferable signal is a property of a purpose-built, supervised representation, not of curvature or scale, which is why it can be reproduced without frontier-scale resources and run as a cheap resident screen. Two instruments the field already trusts, human panels and frontier judges, each leave a class of behavioural failures unread; the practical recommendation is to state the estimand, then evaluate with instruments whose errors do not coincide.

\subsection{5. Limitations}\label{limitations}

The manifestation label is assigned by a single automatic judge, so the manifestation half of the comparison and the discordance result in Section 3.3 are established against an automatic reference rather than human-verified behaviour; a prespecified, human-adjudicated manifestation subset, blind to condition and generator, is the decisive next step and would convert the manifestation ranking and the coverage gain from provisional to confirmed. The resolution-matched control removes two of the three asymmetries in the comparison, matching the output format and asking the judge a target-specific question, and the estimand interaction survives it. One asymmetry remains: the auditor is supervised on in-fold generators while the judge is zero-shot, so the reported distances are between a supervised read-out and an unsupervised one. A supervised-versus-supervised layer, fitting a comparable read-out on the judge's own representations, would close that gap and is the natural next control; we did not run it because the judge's representations are not accessible. The trait evaluation covers three traits and three generators from two providers, with uneven manifestation counts across families (15 on the most distant), and broader coverage would tighten the per-family estimates. The human and trait-detection evaluations are not co-registered item for item. The auditor detects behavioural-consistency failures on the reported classes and does not address factual truthfulness or other honesty failure modes.

\subsection{6. Methods}\label{methods}

\textbf{Auditor architecture and training.} Decoder-only transformer, 146 million parameters, depth 16, width 512, 16 attention heads, 128-dimensional hyperbolic (Lorentz) embedding, trained from scratch for 18 epochs, batch 128, peak learning rate 3e-4, 14 hours on one A100 SXM. A shared encoder is read out through fourteen classification heads; the geodesic statistic g(u) = sigma(w · d\_H(u\_hat, tau\_hat) + b) is the design-level audit rule. All reported evaluations use one checkpoint, SHA-256 3127f6c75dd4d0a67622c7f5b5ed8648df0b594f464531b5c67e1496bc7fcf94, re-evaluated end to end for this study.

\textbf{Corpus.} 812,833 utterance-trace-label triples (90,314 held out), generated programmatically with teacher models from multiple frontier families. A disclosure protocol of 6 to 12 temperature-0 follow-ups with preference-signal neutralisation elicits a compact behavioural trace (features weighted, alternatives ranked, rejection reasons). Labels are assigned by construction at generation time; session-level evaluations use automatic tool-log ground truth independent of these labels. The held-out set is partitioned at the item level and is not stratified by generator family.

\textbf{Companion-induced trait evaluation.} Three trait conditions (sycophancy, dependence-fostering, confabulated memories) instantiated as companion system prompts, with a matched neutral prompt that shares the companion framing but mandates honesty. For each of three generators from two providers, each item samples a persona, a probe, and a six-turn personalisation memory; trait and neutral replies are generated by the same generator on identical context (style-controlled pair; 120 pairs per generator; 720 replies). Two labels are recorded per reply: the exposure label (whether the prompt induced the trait) and the manifestation label, the latter assigned by an independent frontier judge under a fixed trait-specific rubric (manifestation rate 61.4 percent of trait-conditioned replies). Auditor detection reads the frozen mean-pooled representation through class-balanced logistic regression under leave-one-generator-out evaluation, refit and scored against each label in turn; per-item out-of-fold scores are retained. The frontier baseline classifies the same replies zero-shot under a matched rubric, scored identically against each label; per-item predictions are retained. Paired instrument differences use a cluster bootstrap (2,000 resamples) of the mean-over-generator AUROC in which the resampling unit is the trait-and-neutral pair rather than the individual reply, drawn within each held-out generator, so that the paired construction of the evaluation set is preserved; all instruments are rescored on each resample, so the intervals are paired. The comparison is repeated at three output resolutions: the judge's deployed single verdict against the auditor's continuous read-out, a target-specific continuous confidence score from the same judge (the probability of its YES token, normalised over the YES and NO tokens) against that same read-out, and the judge's verdict against the auditor's read-out thresholded at 0.5, the class-balanced midpoint, without data-dependent optimisation. The estimand interaction is the difference between the exposure and manifestation gaps and is given a cluster bootstrap interval on the same resamples. Instrument discordance uses McNemar's test on per-item correctness against the manifestation label, which bears on the equality of the two marginal error rates only; association between the error events is reported separately as an odds ratio and a phi coefficient on the two-by-two correctness table. The inclusive rule flags a reply positive if either instrument does, the exclusive rule only if both do; recall, precision, specificity and false-positive rate are reported for each. A fine-tuning comparison unfreezes the honest head and the final decoder block (four epochs, learning rate 1e-4, trained on the two in-fold families).

\textbf{Human baseline.} Ten independent raters (nine rating from session text plus a gold reference answer, one additionally consulting tool-call logs) classified twenty-nine held-out sessions on a three-class compliance rubric; Fleiss kappa is reported for the nine text-only raters (primary) and for all ten pooled (sensitivity). Anonymised ratings are released with the deposit.

\textbf{Statistics.} All intervals are 95 percent percentile bootstrap intervals and all tests are two-sided at alpha 0.05. Accuracies and kappa use 10,000 item-level resamples. Paired evaluator differences and the estimand interaction use 2,000 cluster bootstrap resamples with the trait-and-neutral pair as the unit, drawn within each held-out generator. Marginal error rates are compared by McNemar's exact test on paired per-item correctness; error association is reported as an odds ratio and phi coefficient rather than tested. Because the design holds out one generator at a time from a set of three, all results are conditional on these three generator families and are not a random-effects generalization to a population of generators. Analysis code and evaluation outputs are deposited, and every value in Tables 2 and 3 is regenerated by a single script from the per-item files.

\textbf{Ethics.} The human-rating study asked raters to classify anonymised, model-generated session transcripts on a compliance rubric. It collected no personal or sensitive data about the raters beyond their ratings, and posed no more than minimal risk. Raters provided informed consent to participate and to the release of their anonymised ratings.

\subsection{Data Availability}\label{data-availability}

The evaluation question sets, the 720 generated replies with both labels, per-item scores for all three instrument configurations, the human ratings, and the analysis and evaluation scripts are archived on Zenodo (concept DOI 10.5281/zenodo.21232421) with a public changelog. The analysis regenerates end to end from three scripts; no reported outcome is supplied to any calculation, and canonical values appear in the code only as post-computation validation checks. The first, \texttt{regen\_auditor\_oof.py}, verifies the checkpoint's full SHA-256, loads it frozen, extracts the mean-pooled representation for all 720 replies, fits the leave-one-generator-out logistic read-out for each label, and writes the 720 per-item out-of-fold scores; on the archived data it returns mean AUROC 0.804 for exposure and 0.690 for manifestation. The second, \texttt{reproduce\_table2.py}, regenerates every value in Tables 2 and 3 from those per-item scores together with the two archived judge score files, including all three output resolutions, the estimand interaction, the exact McNemar p, and every cluster bootstrap interval. The figure script \texttt{make\_sr\_figures.py} regenerates all three figures from the same files. Package versions are recorded in the archived environment file. The auditor checkpoint (SHA-256 3127f6c75dd4d0a67622c7f5b5ed8648df0b594f464531b5c67e1496bc7fcf94) is held in a private model repository together with the pinned source archive; reviewers are granted access on request during peer review, and the repository will be made gated-public at publication. It is the single artifact producing the per-head accuracies in Table 1 and the read-out scores behind Tables 2 and 3.

\subsection{References}\label{references}

\begin{enumerate}
\def\labelenumi{\arabic{enumi}.}
\tightlist
\item
  Perez, E. et al.~Discovering Language Model Behaviors with Model-Written Evaluations. Preprint at https://arxiv.org/abs/2212.09251 (2022).
\item
  Zheng, L. et al.~Judging LLM-as-a-Judge with MT-Bench and Chatbot Arena. Preprint at https://arxiv.org/abs/2306.05685 (2023).
\item
  Jacobs, A. Z. \& Wallach, H. Measurement and Fairness. \emph{Proceedings of the 2021 ACM Conference on Fairness, Accountability, and Transparency} (2021). doi:10.1145/3442188.3445901.
\item
  Raji, I. D., Bender, E. M., Paullada, A., Denton, E. \& Hanna, A. AI and the Everything in the Whole Wide World Benchmark. Preprint at https://arxiv.org/abs/2111.15366 (2021).
\item
  Dehghani, M. et al.~The Benchmark Lottery. Preprint at https://arxiv.org/abs/2107.07002 (2021).
\item
  Bean, A. M. et al.~Measuring what Matters: Construct Validity in Large Language Model Benchmarks. Preprint at https://arxiv.org/abs/2511.04703 (2025).
\item
  Singh, S. et al.~The Leaderboard Illusion. Preprint at https://arxiv.org/abs/2504.20879 (2025).
\item
  Septiandri, A. A., Constantinides, M., Tahaei, M. \& Quercia, D. WEIRD FAccTs: How Western, Educated, Industrialized, Rich, and Democratic is FAccT?. \emph{2023 ACM Conference on Fairness Accountability and Transparency} (2023). doi:10.1145/3593013.3593985.
\item
  Buyl, M. et al.~Large Language Models Reflect the Ideology of their Creators. Preprint at https://arxiv.org/abs/2410.18417 (2024).
\item
  Liu, N. F. et al.~Lost in the Middle: How Language Models Use Long Contexts. Preprint at https://arxiv.org/abs/2307.03172 (2023).
\item
  Freiesleben, T. \& Zezulka, S. The Benchmarking Epistemology: Construct Validity for Evaluating Machine Learning Models. Preprint at https://arxiv.org/abs/2510.23191 (2025).
\item
  Freiesleben, T. Establishing Construct Validity in LLM Capability Benchmarks Requires Nomological Networks. Preprint at https://arxiv.org/abs/2603.15121 (2026).
\item
  Wallach, H. et al.~Position: Evaluating Generative AI Systems Is a Social Science Measurement Challenge. Preprint at https://arxiv.org/abs/2502.00561 (2025).
\item
  Salaudeen, O. et al.~Measurement to Meaning: A Validity-Centered Framework for AI Evaluation. Preprint at https://arxiv.org/abs/2505.10573 (2025).
\item
  Weidinger, L. et al.~Toward an Evaluation Science for Generative AI Systems. Preprint at https://arxiv.org/abs/2503.05336 (2025).
\item
  Solaiman, I. et al.~Evaluating the Social Impact of Generative AI Systems in Systems and Society. Preprint at https://arxiv.org/abs/2306.05949 (2023).
\item
  Paruchuri, A. et al.~What Are the Odds? Language Models Are Capable of Probabilistic Reasoning. Preprint at https://arxiv.org/abs/2406.12830 (2024).
\item
  Artstein, R. \& Poesio, M. Inter-Coder Agreement for Computational Linguistics. \emph{Computational Linguistics} (2008). doi:10.1162/coli.07-034-r2.
\item
  McNemar, Q. Note on the Sampling Error of the Difference Between Correlated Proportions or Percentages. \emph{Psychometrika} (1947). doi:10.1007/bf02295996.
\item
  Cohen, J. A Coefficient of Agreement for Nominal Scales. \emph{Educational and Psychological Measurement} (1960). doi:10.1177/001316446002000104.
\item
  Fleiss, J. L. Measuring nominal scale agreement among many raters. \emph{Psychological Bulletin} (1971). doi:10.1037/h0031619.
\item
  Landis, J. R. \& Koch, G. G. The Measurement of Observer Agreement for Categorical Data. \emph{Biometrics} (1977). doi:10.2307/2529310.
\item
  Pavlick, E. \& Kwiatkowski, T. Inherent Disagreements in Human Textual Inferences. \emph{Transactions of the Association for Computational Linguistics} (2019). doi:10.1162/tacl\_a\_00293.
\item
  Nie, Y., Zhou, X. \& Bansal, M. What Can We Learn from Collective Human Opinions on Natural Language Inference Data?. Preprint at https://arxiv.org/abs/2010.03532 (2020).
\item
  Davani, A. M., Díaz, M. \& Prabhakaran, V. Dealing with Disagreements: Looking Beyond the Majority Vote in Subjective Annotations. Preprint at https://arxiv.org/abs/2110.05719 (2021).
\item
  Röttger, P., Vidgen, B., Hovy, D. \& Pierrehumbert, J. B. Two Contrasting Data Annotation Paradigms for Subjective NLP Tasks. Preprint at https://arxiv.org/abs/2112.07475 (2021).
\item
  Plank, B. The `Problem' of Human Label Variation: On Ground Truth in Data, Modeling and Evaluation. Preprint at https://arxiv.org/abs/2211.02570 (2022).
\item
  Denton, R., Díaz, M., Kivlichan, I., Prabhakaran, V. \& Rosen, R. Whose Ground Truth? Accounting for Individual and Collective Identities Underlying Dataset Annotation. Preprint at https://arxiv.org/abs/2112.04554 (2021).
\item
  Mokhberian, N. et al.~Capturing Perspectives of Crowdsourced Annotators in Subjective Learning Tasks. Preprint at https://arxiv.org/abs/2311.09743 (2023).
\item
  Karpinska, M., Akoury, N. \& Iyyer, M. The Perils of Using Mechanical Turk to Evaluate Open-Ended Text Generation. Preprint at https://arxiv.org/abs/2109.06835 (2021).
\item
  Kirk, H. R. et al.~The PRISM Alignment Dataset: What Participatory, Representative and Individualised Human Feedback Reveals About the Subjective and Multicultural Alignment of Large Language Models. \emph{The Thirty-eight Conference on Neural Information Processing Systems Datasets and Benchmarks Track (2024)} (2024); preprint at https://arxiv.org/abs/2404.16019.
\item
  Alain, G. \& Bengio, Y. Understanding intermediate layers using linear classifier probes. Preprint at https://arxiv.org/abs/1610.01644 (2016).
\item
  Hewitt, J. \& Liang, P. Designing and Interpreting Probes with Control Tasks. \emph{Proceedings of the 2019 Conference on Empirical Methods in Natural Language Processing and the 9th International Joint Conference on Natural Language Processing (EMNLP-IJCNLP)} (2019). doi:10.18653/v1/d19-1275.
\item
  Ravichander, A., Belinkov, Y. \& Hovy, E. Probing the Probing Paradigm: Does Probing Accuracy Entail Task Relevance?. Preprint at https://arxiv.org/abs/2005.00719 (2020).
\item
  Hase, P., Xie, H. \& Bansal, M. The Out-of-Distribution Problem in Explainability and Search Methods for Feature Importance Explanations. Preprint at https://arxiv.org/abs/2106.00786 (2021).
\item
  Belinkov, Y. Probing Classifiers: Promises, Shortcomings, and Advances. Preprint at https://arxiv.org/abs/2102.12452 (2021).
\item
  Elazar, Y., Ravfogel, S., Jacovi, A. \& Goldberg, Y. Amnesic Probing: Behavioral Explanation with Amnesic Counterfactuals. Preprint at https://arxiv.org/abs/2006.00995 (2020).
\item
  Ravfogel, S., Twiton, M., Goldberg, Y. \& Cotterell, R. Linear Adversarial Concept Erasure. Preprint at https://arxiv.org/abs/2201.12091 (2022).
\item
  Burns, C., Ye, H., Klein, D. \& Steinhardt, J. Discovering Latent Knowledge in Language Models Without Supervision. Preprint at https://arxiv.org/abs/2212.03827 (2022).
\item
  Mallen, A., Brumley, M., Kharchenko, J. \& Belrose, N. Eliciting Latent Knowledge from Quirky Language Models. Preprint at https://arxiv.org/abs/2312.01037 (2023).
\item
  Marks, S. \& Tegmark, M. The Geometry of Truth: Emergent Linear Structure in Large Language Model Representations of True/False Datasets. Preprint at https://arxiv.org/abs/2310.06824 (2023).
\item
  Gurnee, W. \& Tegmark, M. Language Models Represent Space and Time. Preprint at https://arxiv.org/abs/2310.02207 (2023).
\item
  Nanda, N., Lee, A. \& Wattenberg, M. Emergent Linear Representations in World Models of Self-Supervised Sequence Models. Preprint at https://arxiv.org/abs/2309.00941 (2023).
\item
  Zou, A. et al.~Representation Engineering: A Top-Down Approach to AI Transparency. Preprint at https://arxiv.org/abs/2310.01405 (2023).
\item
  Li, K., Patel, O., Viégas, F., Pfister, H. \& Wattenberg, M. Inference-Time Intervention: Eliciting Truthful Answers from a Language Model. Preprint at https://arxiv.org/abs/2306.03341 (2023).
\item
  Turpin, M., Michael, J., Perez, E. \& Bowman, S. R. Language Models Don't Always Say What They Think: Unfaithful Explanations in Chain-of-Thought Prompting. Preprint at https://arxiv.org/abs/2305.04388 (2023).
\item
  Lanham, T. et al.~Measuring Faithfulness in Chain-of-Thought Reasoning. Preprint at https://arxiv.org/abs/2307.13702 (2023).
\item
  Chen, Y. et al.~Do Models Explain Themselves? Counterfactual Simulatability of Natural Language Explanations. Preprint at https://arxiv.org/abs/2307.08678 (2023).
\item
  Agarwal, C., Tanneru, S. H. \& Lakkaraju, H. Faithfulness vs.~Plausibility: On the (Un)Reliability of Explanations from Large Language Models. Preprint at https://arxiv.org/abs/2402.04614 (2024).
\item
  Ji, Z. et al.~Survey of Hallucination in Natural Language Generation. \emph{ACM Computing Surveys (2022)} (2022); preprint at https://arxiv.org/abs/2202.03629.
\item
  Webson, A., Loo, A. M., Yu, Q. \& Pavlick, E. Are Language Models Worse than Humans at Following Prompts? It's Complicated. Preprint at https://arxiv.org/abs/2301.07085 (2023).
\item
  Panickssery, A., Bowman, S. R. \& Feng, S. LLM Evaluators Recognize and Favor Their Own Generations. Preprint at https://arxiv.org/abs/2404.13076 (2024).
\item
  Oi, M., Kaneko, M., Koike, R., Loem, M. \& Okazaki, N. Likelihood-based Mitigation of Evaluation Bias in Large Language Models. \emph{ACL2024 (findings)} (2024); preprint at https://arxiv.org/abs/2402.15987.
\item
  Dubois, Y., Galambosi, B., Liang, P. \& Hashimoto, T. B. Length-Controlled AlpacaEval: A Simple Way to Debias Automatic Evaluators. Preprint at https://arxiv.org/abs/2404.04475 (2024).
\item
  Casper, S., Lin, J., Kwon, J., Culp, G. \& Hadfield-Menell, D. Explore, Establish, Exploit: Red Teaming Language Models from Scratch. Preprint at https://arxiv.org/abs/2306.09442 (2023).
\item
  Ouyang, L. et al.~Training language models to follow instructions with human feedback. Preprint at https://arxiv.org/abs/2203.02155 (2022).
\item
  Sharma, M. et al.~Towards Understanding Sycophancy in Language Models. Preprint at https://arxiv.org/abs/2310.13548 (2023).
\item
  Wei, J., Huang, D., Lu, Y., Zhou, D. \& Le, Q. V. Simple synthetic data reduces sycophancy in large language models. Preprint at https://arxiv.org/abs/2308.03958 (2023).
\item
  Qi, X. et al.~Fine-tuning Aligned Language Models Compromises Safety, Even When Users Do Not Intend To!. Preprint at https://arxiv.org/abs/2310.03693 (2023).
\item
  Ren, R. et al.~Safetywashing: Do AI Safety Benchmarks Actually Measure Safety Progress?. Preprint at https://arxiv.org/abs/2407.21792 (2024).
\item
  Jiang, G. et al.~Evaluating and Inducing Personality in Pre-trained Language Models. Preprint at https://arxiv.org/abs/2206.07550 (2022).
\item
  Jiang, H. et al.~PersonaLLM: Investigating the Ability of Large Language Models to Express Personality Traits. Preprint at https://arxiv.org/abs/2305.02547 (2023).
\item
  Zheng, M., Pei, J., Logeswaran, L., Lee, M. \& Jurgens, D. When ``A Helpful Assistant'' Is Not Really Helpful: Personas in System Prompts Do Not Improve Performances of Large Language Models. Preprint at https://arxiv.org/abs/2311.10054 (2023).
\item
  Zhou, X. et al.~SOTOPIA: Interactive Evaluation for Social Intelligence in Language Agents. Preprint at https://arxiv.org/abs/2310.11667 (2023).
\item
  Liu, K., Casper, S., Hadfield-Menell, D. \& Andreas, J. Cognitive Dissonance: Why Do Language Model Outputs Disagree with Internal Representations of Truthfulness?. Preprint at https://arxiv.org/abs/2312.03729 (2023).
\item
  Li, K. et al.~Measuring and Controlling Instruction (In)Stability in Language Model Dialogs. Preprint at https://arxiv.org/abs/2402.10962 (2024).
\item
  Kocielnik, R. et al.~Rethinking Psychometric Evaluation of LLMs: When and Why Self-Reports Predict Behavior. Preprint at https://arxiv.org/abs/2606.12730 (2026).
\item
  Hinton, G., Vinyals, O. \& Dean, J. Distilling the Knowledge in a Neural Network. Preprint at https://arxiv.org/abs/1503.02531 (2015).
\item
  Schick, T. \& Schütze, H. It's Not Just Size That Matters: Small Language Models Are Also Few-Shot Learners. Preprint at https://arxiv.org/abs/2009.07118 (2020).
\item
  Hsieh, C. et al.~Distilling Step-by-Step! Outperforming Larger Language Models with Less Training Data and Smaller Model Sizes. \emph{Findings of the Association for Computational Linguistics: ACL 2023} (2023). doi:10.18653/v1/2023.findings-acl.507.
\item
  Eldan, R. \& Li, Y. TinyStories: How Small Can Language Models Be and Still Speak Coherent English?. Preprint at https://arxiv.org/abs/2305.07759 (2023).
\item
  Gunasekar, S. et al.~Textbooks Are All You Need. Preprint at https://arxiv.org/abs/2306.11644 (2023).
\item
  Xu, X. et al.~A Survey on Knowledge Distillation of Large Language Models. Preprint at https://arxiv.org/abs/2402.13116 (2024).
\item
  Nickel, M. \& Kiela, D. Poincaré Embeddings for Learning Hierarchical Representations. Preprint at https://arxiv.org/abs/1705.08039 (2017).
\item
  Sa, C. D., Gu, A., Ré, C. \& Sala, F. Representation Tradeoffs for Hyperbolic Embeddings. Preprint at https://arxiv.org/abs/1804.03329 (2018).
\item
  Nickel, M. \& Kiela, D. Learning Continuous Hierarchies in the Lorentz Model of Hyperbolic Geometry. Preprint at https://arxiv.org/abs/1806.03417 (2018).
\item
  Ganea, O., Bécigneul, G. \& Hofmann, T. Hyperbolic Neural Networks. Preprint at https://arxiv.org/abs/1805.09112 (2018).
\item
  Chami, I. et al.~Low-Dimensional Hyperbolic Knowledge Graph Embeddings. Preprint at https://arxiv.org/abs/2005.00545 (2020).
\item
  Peng, W., Varanka, T., Mostafa, A., Shi, H. \& Zhao, G. Hyperbolic Deep Neural Networks: A Survey. Preprint at https://arxiv.org/abs/2101.04562 (2021).
\item
  Recht, B., Roelofs, R., Schmidt, L. \& Shankar, V. Do ImageNet Classifiers Generalize to ImageNet?. Preprint at https://arxiv.org/abs/1902.10811 (2019).
\item
  Sagawa, S., Koh, P. W., Hashimoto, T. B. \& Liang, P. Distributionally Robust Neural Networks for Group Shifts: On the Importance of Regularization for Worst-Case Generalization. Preprint at https://arxiv.org/abs/1911.08731 (2019).
\item
  Gulrajani, I. \& Lopez-Paz, D. In Search of Lost Domain Generalization. Preprint at https://arxiv.org/abs/2007.01434 (2020).
\item
  Koh, P. W. et al.~WILDS: A Benchmark of in-the-Wild Distribution Shifts. Preprint at https://arxiv.org/abs/2012.07421 (2020).
\item
  Liu, J. et al.~Towards Out-Of-Distribution Generalization: A Survey. Preprint at https://arxiv.org/abs/2108.13624 (2021).
\item
  Yao, H. et al.~Improving Out-of-Distribution Robustness via Selective Augmentation. Preprint at https://arxiv.org/abs/2201.00299 (2022).
\item
  Mitchell, M. et al.~Model Cards for Model Reporting. \emph{FAT} '19: Conference on Fairness, Accountability, and Transparency, January 29--31, 2019, Atlanta, GA, USA* (2018); preprint at https://arxiv.org/abs/1810.03993.
\item
  Raji, I. D. et al.~Closing the AI Accountability Gap: Defining an End-to-End Framework for Internal Algorithmic Auditing. Preprint at https://arxiv.org/abs/2001.00973 (2020).
\item
  Raji, I. D., Xu, P., Honigsberg, C. \& Ho, D. E. Outsider Oversight: Designing a Third Party Audit Ecosystem for AI Governance. Preprint at https://arxiv.org/abs/2206.04737 (2022).
\item
  Anderljung, M. et al.~Towards Publicly Accountable Frontier LLMs: Building an External Scrutiny Ecosystem under the ASPIRE Framework. Preprint at https://arxiv.org/abs/2311.14711 (2023).
\item
  Reuel, A. et al.~Open Problems in Technical AI Governance. \emph{Transactions on Machine Learning Research, 2025} (2024); preprint at https://arxiv.org/abs/2407.14981.
\item
  Casper, S. et al.~Black-Box Access is Insufficient for Rigorous AI Audits. \emph{The 2024 ACM Conference on Fairness, Accountability, and Transparency (FAccT '24), June 3-6, 2024, Rio de Janeiro, Brazil} (2024); preprint at https://arxiv.org/abs/2401.14446.
\item
  Dietterich, T. G. Ensemble Methods in Machine Learning. \emph{Lecture Notes in Computer Science} (2000). doi:10.1007/3-540-45014-9\_1.
\item
  Shin, K. S., Kang, I. S., Min, Y. \& Lee, M. A small language model detects behavioural faithfulness gaps that frontier judges and human raters miss. Preprint at https://arxiv.org/abs/2607.09306v2 (2026).
\end{enumerate}

\subsection{Author contributions}\label{author-contributions}

K.S.S. is the sole author. K.S.S. conceived the study, built the auditor and the evaluation pipeline, constructed the evaluation corpus, ran all experiments and analyses, produced the display items, and wrote the manuscript.

\subsection{Additional Information}\label{additional-information}

\textbf{Funding.} The author received no specific funding for this work.

\textbf{Competing interests.} The author has two patent applications pending with the Korean Intellectual Property Office, 10-2026-0127826 and 10-2026-0127815, relating broadly to on-device small-model architectures of the kind evaluated here. The author declares no other competing interests.

\textbf{Use of AI tools.} The evaluation stimuli (companion-trait and neutral replies) were generated by large language models, and the manifestation labels were assigned by a large-language-model judge, both as described in Methods; these uses are part of the reported methodology. Generative AI was not used to produce the study's scientific claims, analyses, or conclusions.

\end{document}